\pdfoutput=1
\documentclass{article}

    \PassOptionsToPackage{sort, square, numbers}{natbib}

     \usepackage[preprint]{neurips_2020}

\usepackage[utf8]{inputenc} 
\usepackage[T1]{fontenc}    
\usepackage{hyperref}       
\usepackage{url}            
\usepackage{booktabs}       
\usepackage{amsfonts}       
\usepackage{nicefrac}       
\usepackage{microtype}      

\usepackage{xcolor}
\usepackage{graphicx}
\usepackage{subcaption}
\usepackage[export]{adjustbox}
\usepackage{amsmath}
\usepackage{wrapfig}
\usepackage[ruled,vlined]{algorithm2e}
\usepackage{color}

\newcommand{\newparagraph}[1]{\noindent\textbf{#1\hspace{1em}}}

\title{Principal Component Networks: \\Parameter Reduction Early in Training}

\author{
  Roger Waleffe \\
  Department of Computer Science\\
  University of Wisconsin-Madison\\
  \texttt{waleffe@wisc.edu} \\
  \And
  Theodoros Rekatsinas \\
  Department of Computer Science \\
  University of Wisconsin-Madison \\
  \texttt{thodrek@cs.wisc.edu} \\
}

\begin{document}

\maketitle

\begin{abstract}
Recent works show that overparameterized networks contain small subnetworks that exhibit comparable accuracy to the full model when trained in isolation. These results highlight the potential to reduce training costs of deep neural networks without sacrificing generalization performance. However, existing approaches for finding these small networks rely on expensive multi-round train-and-prune procedures and are non-practical for large data sets and models. In this paper, we show how to find small networks that exhibit the same performance as their overparameterized counterparts after only a few training epochs. We find that hidden layer activations in overparameterized networks exist primarily in subspaces smaller than the actual model width. Building on this observation, we use PCA to find a basis of high variance for layer inputs and represent layer weights using these directions. We eliminate all weights not relevant to the found PCA basis and term these network architectures Principal Component Networks. On CIFAR-10 and ImageNet, we show that PCNs train faster and use less energy than overparameterized models, without accuracy loss. We find that our transformation leads to networks with up to 23.8$\times$ fewer parameters, with equal or higher end-model accuracy---in some cases we observe improvements up to 3\%. We also show that ResNet-20 PCNs outperform deep ResNet-110 networks while training faster.
\end{abstract}

\section{Introduction}
\label{intro}
Recent results suggest the importance of overparameterization in neural networks~\cite{gunasekar_2017, li_2020}. The theoretical results of~\citet{gunasekar_2017} demonstrate that training in an overparameterized regime leads to an implicit regularization that may improve generalization. At the same time, empirical results~\cite{li_2020} show that large models can lead to higher test accuracy. Yet, these generalization improvements come with increased time and resource utilization costs: models with more parameters generally require more FLOPS. The question then arises, \emph{how can we retain the generalization benefits of overparameterized training but reduce its computational cost?}

This question has motivated several works which show that overparameterized networks contain \emph{sparse subnetworks} with comparable generalization performance~\cite{frankle_2019, ramanujan_2019, blalock_2020}. 
Currently, accurate sparse models are obtained either by pruning the weights of a fully trained network~\cite{blalock_2020},---this approach targets efficient inference---or by using iterative train-prune-reset procedures during model learning~\cite{frankle_2019, ramanujan_2019}. In particular, subnetworks discovered by~\citet{frankle_2019} are capable of training in isolation without accuracy loss---the so called \emph{lottery ticket hypothesis}. Subnetwork identification, however, can be an expensive exercise, limiting the potential to reduce end-to-end training costs. As pointed out by~\citet{frankle_2019}, train-time iterative pruning cannot be effectively applied to large data sets and models such as ImageNet due to its computational cost. This motivates our study to find an efficient procedure for discovering small networks (during training) that exhibit the same performance as their overparameterized counterparts.

\newparagraph{Principal Component Networks}
We show that after only a few training epochs, overparameterized networks can be transformed into small networks that exhibit comparable generalization performance. Our work builds upon the next compelling observation: We consider overparameterization due to increased network width, a key quantity associated with high accuracy models~\cite{chen_2018, park_2019}. We find that hidden layer activations in wide networks exist in low-dimensional subspaces an order of magnitude smaller than the actual model width. We also find that these subspaces, which contribute most to the generalization accuracy of the network, can be identified early in training.

Based on the above observation, we introduce a new family of deep learning models, which we term \emph{Principal Component Networks} (PCNs). A PCN transforms the wide layers in an original overparameterized network into \emph{smaller layers} that live in a lower dimensional space. To identify the basis of this space for each layer, we use Principal Component Analysis (PCA) to find the high variance directions that describe the layer’s input and output activations. The transformations introduced by PCNs eliminate all weights from the overparameterized model not relevant to these bases. In summary, PCNs introduce the following procedure for identifying and training small networks that do not sacrifice end-model accuracy:
\begin{enumerate}
    \item Randomly initialize a wide, overparameterized neural network.
    \item Train the network for a few epochs  (in some cases as little as one epoch).
    \item Use PCA to find the low-dimensional spaces of network activations and project the weights of the network onto these subspaces to obtain an equivalent PCN.
    \item Continue training the PCN until convergence.
\end{enumerate}
This procedure can be applied to a variety of neural networks, including dense neural networks (DNNs), convolutional neural networks (CNNs), and residual neural networks (ResNets).

\newparagraph{Results}
We empirically validate training PCNs on CIFAR-10~\cite{krizhevsky_2009} and ImageNet~\cite{russakovsky_2015} and compare against training the corresponding overparameterized model. For both ResNet~\cite{he_2016} and VGG-style architectures~\cite{simonyan_2014}, we show that PCNs train faster and use less energy, without sacrificing end-model accuracy. Our transformed PCNs have up to 23.8$\times$ fewer parameters and in some cases exhibit accuracy improvements of up to 3\%. Interestingly, we show that PCNs derived from wide ResNet models have less parameters but achieve higher accuracy than deep ResNet architectures. Our wide ResNet-20 PCN outperforms a deep ResNet-110 by 0.58\% while training 49\% faster per epoch.

\section{Background and motivation}
\label{motivation}
We first review PCA and then present the empirical observations that motivate our work. 

\begin{wrapfigure}{r}{0.4\linewidth}
    \vspace{-12pt}
    \centering
    \includegraphics[width=0.22\textwidth]{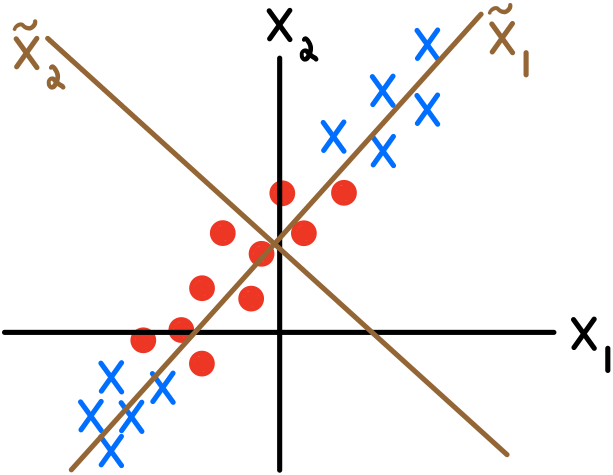}
    \caption{PCA illustration with principal components shown in brown.}
    \label{fig:pca_illustration}
\end{wrapfigure}
\newparagraph{Principal Component Analysis}
Given a set of $m$-dimensional vectors, PCA computes a new basis for the vector space with the following property: 
The first basis vector (termed Principal Component) is the direction of highest variance among the data. The second basis vector has highest variance among directions perpendicular to the first, and so on. An example is shown in Figure~\ref{fig:pca_illustration}. Two-dimensional vectors in the original $[x_1, x_2]$ basis can also be represented in the $[\tilde{x}_1, \tilde{x}_2]$ PCA basis.
We define the \emph{effective dimensionality $m_e$} of the $m$-dimensional space as the number of PCA directions with variance greater than a threshold $\tau$. In Figure~\ref{fig:pca_illustration}, the spread along $\tilde{x}_2$ does not help differentiate between the two sets of points, as the original two-dimensional vectors exist primarily in a one-dimensional space given by the coordinate $\tilde{x}_1$.

We compute the principal components and associated variances using the spectral decomposition of the covariance matrix.\footnote{We find this method to be significantly faster than using SVD in TensorFlow.} Given $N$ data points in $\mathbb{R}^m$ with empirical mean $\boldsymbol{\mu}_m \in \mathbb{R}^m$ we have:
\begin{equation}
    \label{eqn:pca}
    \mathbf{e}_m, V_{m \times m} = eigh\left( \frac{1}{N-1}\left(X_{N \times m} - \boldsymbol{\mu}_m \right)^T \left(X_{N \times m} - \boldsymbol{\mu}_m \right) \right).
\end{equation}
Function $eigh$ returns two quantities: the vector of eigenvalues (variances) $\mathbf{e}_m$, and a matrix $V_{m \times m}$ whose columns contain the eigenvectors (principal components). 
We assume that both outputs are sorted in descending eigenvalue order. 
To transform a vector $\mathbf{x}_m$ into the PCA basis, one computes: $\tilde{\mathbf{x}}_m=(\mathbf{x}_m-\boldsymbol{\mu}_m)V_{m \times m}$. Subtracting off the mean vector $\boldsymbol{\mu}_m$ centers the PCA coordinate system, ensuring each principal component has mean zero when average over the $N$ data points.

\begin{figure}
    \centering
    \begin{subfigure}[t]{.475\textwidth}
        \includegraphics[width=\textwidth]{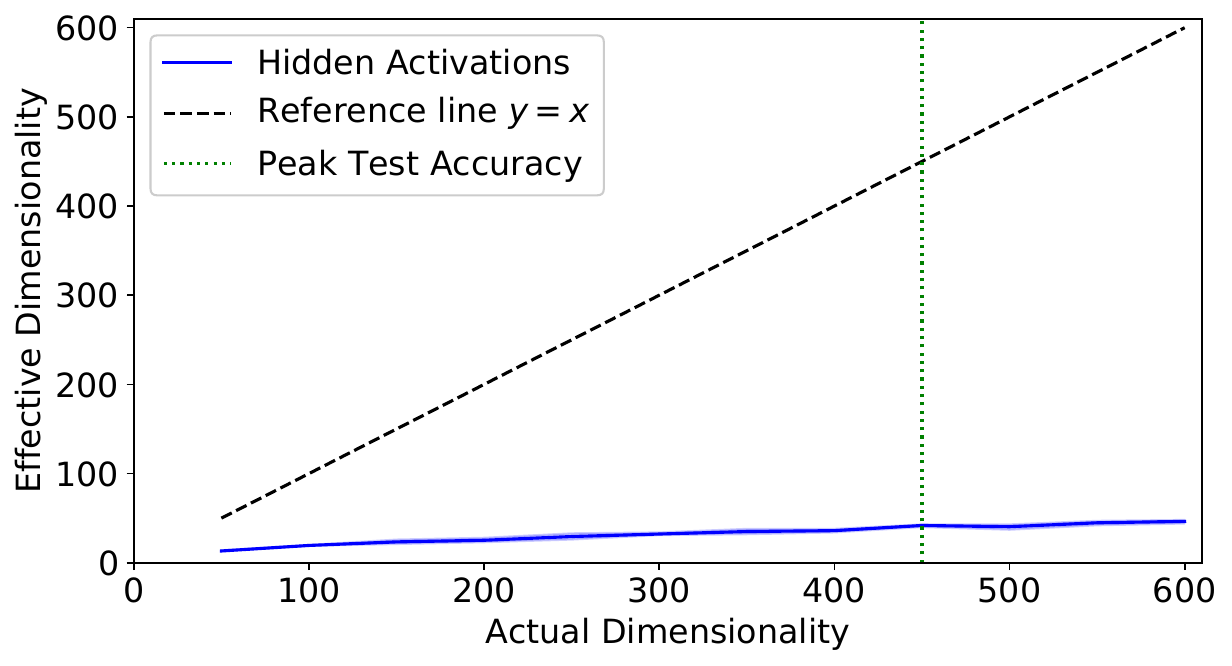}
        \caption{Effective dimensionality vs. width}
        \label{fig:experiment_1}
    \end{subfigure}
    \hspace{.025\textwidth}
    \begin{subfigure}[t]{.475\textwidth}
        \includegraphics[width=\textwidth]{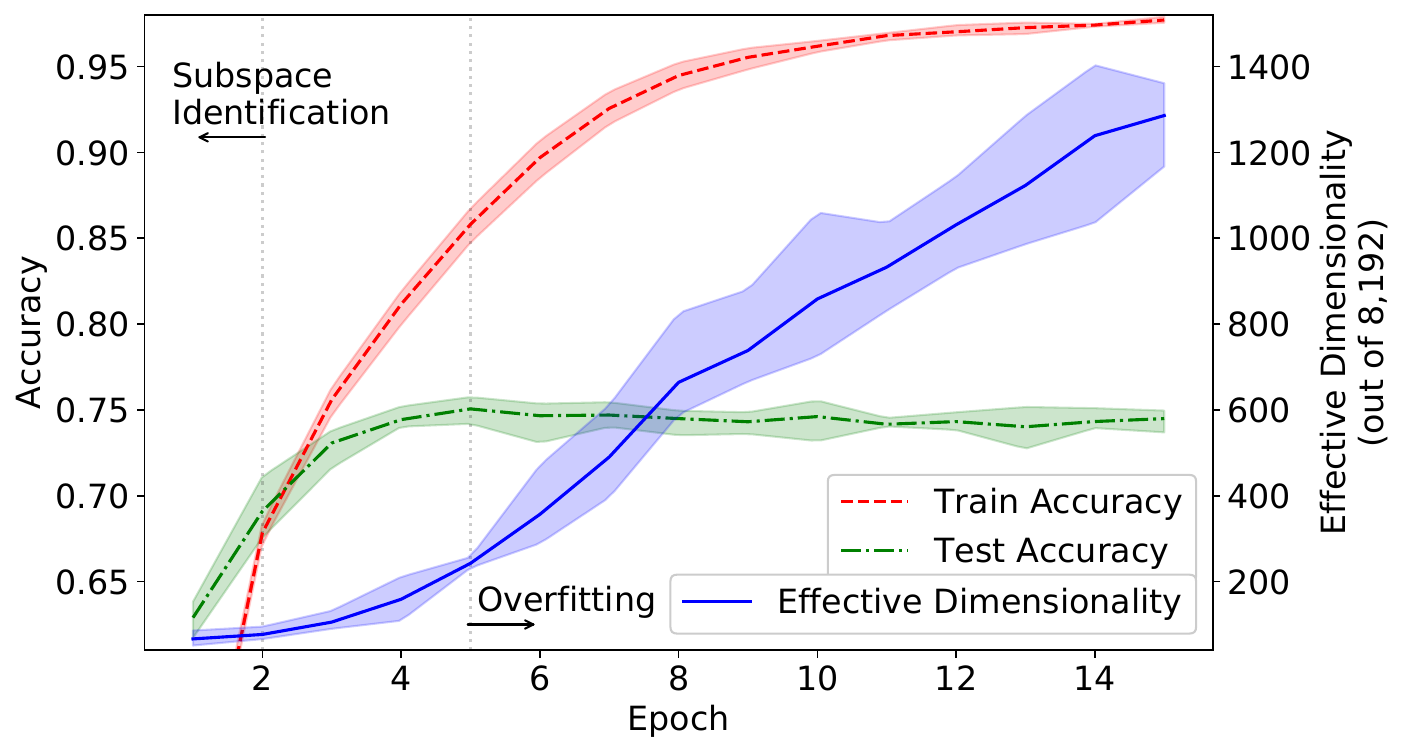}
        \caption{Effective dimensionality vs. training}
        \label{fig:experiment_2}
    \end{subfigure}
    \caption{Effective dimensionality (\# of high-variance activations) for a neural network.}
    \label{fig:motivation_plots}
\end{figure}

\newparagraph{Motivating observations}
We used PCA to measure the effective dimensionality (effective width) of hidden layer activations in neural networks. We present two experiments that highlight our findings.

\textbf{\emph{Experiment 1}} We first analyze the activations of a DNN after learning is complete. We train a DNN with an input layer, a hidden layer with a variable number of nodes, and an output layer with 10 neurons on MNIST~\cite{lecun_1998}. Both dense layers use sigmoid activation and we train until convergence of the validation accuracy. We computed PCA on the network activations \emph{after} the hidden layer and \emph{before} the output layer. The dimension of these activations is equal to the number of nodes in the hidden layer and is thus varied by changing the number of hidden neurons. In Figure \ref{fig:experiment_1} we plot the number of effective dimensions in the activation space versus the number of full dimensions using a PCA variance threshold of 0.1. We see that that the hidden layer activations exist in a subspace with smaller dimension than the full space: with 450 nodes in the hidden layer the network achieves peak test accuracy of 98\%, but rather than occupy 450 dimensions, the hidden layer outputs occupy a space with just over 40 dimensions---a space more than 10$\times$ smaller.

\textbf{\emph{Experiment 2}} We also study the evolution of hidden layer subspaces during training. In this experiment we do not vary the network architecture, and thus, can train a more realistic network. We consider a CNN over CIFAR-10. We use the Conv4 network~\cite{frankle_2019} which consists of four convolution layers followed by three dense layers. After each training epoch, we perform PCA on the hidden activations which form the input to the first fully connected layer. This activation space contains 8,192 dimensions. We calculate PCA for this layer because it contains more than 86\% of the total weights, and thus, affects the end-model accuracy. We \emph{do not} stop training at peak validation accuracy to observe the effective dimensionality of the activation space during overfitting. 

The results are shown in Figure \ref{fig:experiment_2}. We again use a PCA variance threshold of 0.1. There are three regions of interest in this plot: In the first section, up to epoch two, notice that the effective dimensionality grows slowly, \textit{but} the test accuracy grows to 70\%---93\% of its peak at 75\%. The data variance in this $\approx 50$-dimensional subspace is critical for generalization. Contrast this with region three, after the test accuracy peaks at epoch five. Now the network is heavily overfitting, increasing the train accuracy to no avail. Also observe that the network is rapidly creating directions with variance above the 0.1 threshold. All such directions are overfitting to meaningless noise. In between, during epochs 2-5, the effective dimensionality increases from $\approx 50$ to $\approx 200$. We conjecture that these directions are mainly noise, as the test accuracy grows slowly while the train accuracy rapidly surpasses it. The test accuracy can still increase due to \textit{fine-tuning} weights related to the high variance directions discovered in the first region.

\newparagraph{Takeaway}
Experiments 1 and 2 show that hidden layer activations do not occupy their full available dimensions. Rather, they exist in low-dimensional subspaces both during training and once test accuracy peaks. Further, Experiment 2 shows that the few high variance directions which contribute most to generalization accuracy are identified early in training. We use these observations as motivation to transform activations into their high variance PCA bases after only a few epochs.

\section{Principal Component Networks}
\label{pcns}
We now introduce PCNs. First, we describe their application to dense layers. We then extend to CNNs and ResNets. Finally, we discuss the end-to-end training procedure for PCNs. 

\newparagraph{PCNs for dense layers}
Consider a network where the $i^{th}$ hidden layer computes\\ $\mathbf{h}^{i+1}_n = \sigma (\mathbf{h}^{i}_m W^{i}_{m \times n}  + \mathbf{b}^{i}_n)$. Here, $\sigma$ is the activation function, $\mathbf{h}^{i}_m \in \mathbb{R}^m$ is the input activation vector, and $\mathbf{h}^{i+1}_n \in \mathbb{R}^n$ is the output activation vector. $W^{i}_{m \times n}$ and $\mathbf{b}^{i}_n$ are the layer weights and bias vector respectively.
We use superscripts to denote layer index and subscripts to denote dimension when defining a new variable, or for clarity.
Our goal is to transform $W^{i}$ and $\mathbf{b}^{i}$, for each layer $i$, by considering the high variance PCA bases of both the input and output activation spaces.
 
\newparagraph{Transformation based on input}
To transform a layer based on the input activation space, we compute PCA on a batch of input vectors $\mathbf{h}^i$ using Equation \ref{eqn:pca}. Through this calculation, we obtain a mean vector $\boldsymbol{\mu}^i_m$, a vector $\mathbf{e}^i_m$ of variances, and a matrix $V^i_{m \times m}$ whose columns are the principal components (see Section~\ref{motivation}). After finding the PCA basis, we can rewrite any input vector $\mathbf{h}^i$ using these coordinates. If we do so, however, we must also rewrite the layer weight matrix and bias vector. Transformations of $\mathbf{h}^i$, $W^{i}$, and $\mathbf{b}^{i}$ into the PCA basis are given in Equation \ref{eqn:input_identity}. We denote variables represented using PCA coordinates with tilde.  
\begin{equation}
    \label{eqn:input_identity}
    \tilde{\mathbf{h}}^i_m = (\mathbf{h}^i_m - \boldsymbol{\mu}^i_m)V^i_{m \times m}
    \hspace{20pt}
    \tilde{W}^i_{m \times n} = (V^i_{m \times m})^T W^i_{m \times n}
    \hspace{20pt}
    \tilde{\mathbf{b}}^i_n = \mathbf{b}^i_n + \boldsymbol{\mu}^i_m W^i_{m \times n}
\end{equation}
By plugging in the definitions, we have that $\sigma (\tilde{\mathbf{h}}^{i} \tilde{W}^{i} + \tilde{\mathbf{b}}^{i})$ is equivalent to the original $\sigma (\mathbf{h}^{i} W^{i} + \mathbf{b}^{i})$. Changing basis does not change the output.

Instead of performing an identity transformation by using the full PCA basis, we can approximate the original hidden layer by using only the high variance subspace. We define this subspace to contain $m_e$ dimensions, which corresponds to the effective dimensionality of the PCA space under some threshold $\tau$ (see Section \ref{motivation}). Since the columns of $V^i$ are sorted according to decreasing variance, the first $m_e$ columns contain the PCA directions which describe the subspace. We denote the matrix $V^i$ truncated after $m_e$ columns by the matrix $U^i_{m \times m_e}$. In Equation \ref{eqn:input_approx} we rewrite $\mathbf{h}^i$, $W^{i}$, and $\mathbf{b}^{i}$ in the high variance subspace of the input activations.
\begin{equation}
    \label{eqn:input_approx}
    \tilde{\mathbf{h}}^i_{m_e} = (\mathbf{h}^i_m - \boldsymbol{\mu}^i_m)U^i_{m \times m_e}
    \hspace{20pt}
    \tilde{W}^i_{m_e \times n} = (U^i_{m \times m_e})^T W^i_{m \times n}
    \hspace{20pt}
    \tilde{\mathbf{b}}^i_n = \mathbf{b}^i_n + \boldsymbol{\mu}^i_m W^i_{m \times n}
\end{equation}
The hidden layer $\sigma (\tilde{\mathbf{h}}^{i} \tilde{W}^{i} + \tilde{\mathbf{b}}^{i})$ is no longer identical to the original  $\sigma (\mathbf{h}^{i} W^{i}  + \mathbf{b}^{i})$ but it contains fewer weights; The size of the weight matrix reduces from $m \times n$ to $m_e \times n$. We empirically find that often $m_e \ll m$ (see Section~\ref{experiments}). Despite approximating the original hidden layer, we have not changed the dimension of the output $\mathbf{h}^{i+1}$. This means we do not need to modify layer $i+1$.

This approximation introduces limited error to the output of the transformed layer. Errors arise due to dropping the dot product between the last $m-m_e$ elements of $\tilde{\mathbf{h}}^{i}_{m}$ and the last $m-m_e$ rows of $\tilde{W}^{i}_{m \times n}$. However, since $\tilde{\mathbf{h}}^{i}_{m}$ is represented using the PCA basis, the dropped coordinates have mean zero and variance less than the threshold $\tau$. For small $\tau$, the approximate transformation ignores dot products between vectors with $L_2$ norm close to zero and the final $m-m_e$ rows of $\tilde{W}^{i}_{m \times n}$.

\newparagraph{Transformation based on output}
We can also reduce the parameters in layer $i$ using the PCA basis of the subsequent layer $i+1$.
Assume that layer $i$ is transformed per Equation \ref{eqn:input_approx}. Layer $i$ then computes an approximate $\hat{\mathbf{h}}^{i+1}$ output vector: $\hat{\mathbf{h}}^{i+1}_{n} = \sigma (\tilde{\mathbf{h}}^{i}_{m_e} \tilde{W}^{i}_{m_e \times n} + \tilde{\mathbf{b}}^{i}_n)$. We can compress $\tilde{W}^{i}$ and $\tilde{\mathbf{b}}^{i}$ using information about the output activation space. 

Since neural networks compose layers, the next layer $i+1$ will consider $\hat{\mathbf{h}}^{i+1}$ as its \textit{input}. If we consider the input-based transformation for layer $i+1$, we have that $\tilde{\mathbf{h}}^{i+1}_{n_e} = (\hat{\mathbf{h}}^{i+1}_n - \boldsymbol{\mu}^{i+1}_n)U^{i+1}_{n \times n_e}$. We next rewrite $\tilde{\mathbf{h}}^{i+1}$ as a function of the approximate weight matrix $\tilde{W}^{i}$ of the previous layer. We have for the $j^{th}$ component of $\tilde{\mathbf{h}}^{i+1}$:

\begin{align}
    \tilde{\mathbf{h}}^{i+1}[j] & = \sum_{l=1}^n \left(\hat{\mathbf{h}}^{i+1}[l] - \boldsymbol{\mu}^{i+1}[l] \right) U^{i+1}[l, j] \nonumber
    \\ & = \sum_{l=1}^n \left(\sigma\left(\sum_{k=1}^{m_e}\tilde{\mathbf{h}}^{i}[k] \tilde{W}^{i}[k, l] + \tilde{\mathbf{b}}^{i}[l] \right) - \boldsymbol{\mu}^{i+1}[l] \right) U^{i+1}[l, j].
    \label{eqn:output_expansion}
\end{align}
The elements of $\tilde{\mathbf{h}}^{i+1}$ are linear combinations of the centered elements in $\hat{\mathbf{h}}^{i+1}$ weighted by the columns of $U^{i+1}$.
For all elements $j$, the influence of the $l^{th}$ output of layer $i$, denoted $\hat{\mathbf{h}}^{i+1}[l]$, on $\tilde{\mathbf{h}}^{i+1}[j]$ is determined by $U^{i+1}[l, j]$. 
If most entries in the the $l^{th}$ row of $U^{i+1}$ have large values, output $\hat{\mathbf{h}}^{i+1}[l]$ influences many entries in $\tilde{\mathbf{h}}^{i+1}$. On the other hand, if every entry in the $l^{th}$ row of $U^{i+1}$ is small, the $l^{th}$ output of layer $i$ does not influence \textit{any} entry in $\tilde{\mathbf{h}}^{i+1}$. We use the $L_1$ norm of row $l$ in $U^{i+1}$ to determine how important output $\hat{\mathbf{h}}^{i+1}[l]$ is for calculating $\tilde{\mathbf{h}}^{i+1}$. 

We use the above influence measurement to define the output-based transformation of layer $i$. Using the $L_1$ norm criterion described above, we find the subset $S \subseteq [1\dots n]$ of the indices in $\hat{\mathbf{h}}^{i+1}$ with the highest row-wise $L_1$ norm in $U^{i+1}$. The size of $S$ is configurable and can be fixed by the user or determined using an $L_1$-norm threshold. Given $S$, Equation \ref{eqn:output_expansion} becomes:
\begin{equation}
    \label{eqn:output_transform}
    \tilde{\mathbf{h}}^{i+1}[j] = \sum_{l \in S} \left(\sigma\left(\sum_{k=1}^{m_e}\tilde{\mathbf{h}}^{i}[k] \tilde{W}^{i}[k, l] + \tilde{\mathbf{b}}^{i}[l] \right) - \boldsymbol{\mu}^{i+1}[l] \right) U^{i+1}[l, j].
\end{equation}
The columns of $\tilde{W}^{i}$, entries of $\tilde{\mathbf{b}}^{i}$, entries of $\boldsymbol{\mu}^{i+1}$, and rows of $U^{i+1}$ with indices in $S^C$ can be removed from the network. The rows in $W^{i+1}$ with indices in $S^C$ must also be removed so dimensions match when multiplying by $(U^{i+1})^T$ in Equation~\ref{eqn:input_approx}.

The output-based transformation and input-based transformation can be applied to layer $i$ in either order. In the supplementary material we describe the reverse order from the above. 

\newparagraph{PCNs for convolutional layers and ResNets}
We extend the PCN transformations to CNNs and ResNets. We provide a high-level description and leave the details to the supplementary material.  

\newparagraph{Convolutional layers}
To transform convolutional layers, the intuition is the same as for dense layers: Rather than representing hidden layer activations using their default filters, we would like to find a new basis of ``principal filters'' which exhibit sorted high to low variance. 

Convolutional layers operate over matrices and tensors. We denote a convolutional layer\\ $H^{i+1}_{h' \times w' \times n} = \sigma (H^{i}_{h \times w \times m} * W^{i}_{k_1 \times k_2 \times m \times n} + \mathbf{b}^{i}_n)$, where $h$ and $w$ describe the size of the input data, $m$ is the number of input filters, $k_1 \times k_2$ is the kernel size, and $n$ is the number of output filters. 
PCA does not immediately extend to multi-dimensional inputs such as images. As a surrogate, for a batch of $N$ $H_{h \times w \times m}$ matrices, we consider all $h \times w$ depth vectors of dimension $m$ in each image to be examples of points in an $m$ dimensional space. View each axis in this space to be a filter and the point cloud a distribution of how likely each filter is to exhibit a certain pixel value, regardless of $[h, w]$ location. We can then run standard PCA on the flattened set of images $H_{N' \times m}$ with $N' = N \times h \times w$. The resulting $V_{m \times m}$ tells us how to combine \textit{every} depth vector from an original image into a depth vector in the ``principal image''. 

Once we calculate PCA for the input and output activation spaces, i.e., we calculate $V^{i}$ and $V^{i+1}$, we can perform the input- and output-based transformation to convolutional layer $i$. Doing so requires extending Equation \ref{eqn:input_approx} and Equation \ref{eqn:output_transform} to handle tensors instead of vectors and matrices.

\newparagraph{ResNets}
While ResNets consist of convolution layers, they require special care due to the inclusion of residual connections. Since the input transformation (for both dense and convolution layers) modifies only layer $i$ and not subsequent layers, it can be applied immediately to ResNet architectures. The output transformation, however, needs to be modified for some layers. The output of the final layer in a residual block is added to the output of all other residual blocks in a residual stage. Thus, we introduce the added constraint that all layers whose outputs are added together need to perform the output transformation in the same way.

\newparagraph{Training PCNs}
Given a set of layer transformations, we present the training procedure for PCNs.

\newparagraph{Input}
An overparameterized network architecture $N$ with layers $L$, a set of layers $I \subseteq L$ to transform via the input-based transformation, a set of layers $O \subseteq L$ to transform via the output-based transformation, a number of epochs $K$ to train before applying the transformations, a number of epochs $T$ to train after transformation, and a dictionary $C$ containing the thresholds/number of dimensions to use when transforming each layer in $I$ or $O$. Since the output transformation at layer $i$ requires that layer $i+1$ perform the input transformation, there exists a constraint between the sets $I$ and $O$: if $i$ is in $O$ then $i+1$ must be in $I$. Picking the optimal value for $K$ is a challenging problem. Heuristically, we train $N$ so long as the validation accuracy is increasing at a high-rate.

We now describe the steps of the training procedure:

\newparagraph{Step 1}
Train $N$ for $K$ epochs. This step allows us to retain the benefits of the many random initializations present in overparameterized networks and thus achieve high end-model accuracy.

\newparagraph{Step 2}
Use PCA to calculate $\mathbf{e}^i$, $V^i$, and $\boldsymbol{\mu}^i$ for each layer $i \in I$; truncate $V^i$ into $U^i$ using the variances $\mathbf{e}^i$ and $C[i]$, the compression configuration for layer $i$.

Then, transform layers from $N$ into their PCN versions. 
 
\newparagraph{Step 3}
For each layer $i \in O$ we perform the output-based transformation.

\newparagraph{Step 4}
Transform all layers $i \in I$ using the input-based transformation.

\newparagraph{Step 5}
Finally, create the PCN and train for $T$ epochs. Here, we do not update the $U$ matrices. The trainable parameters are only the layer weight matrices and bias vectors.

\section{Experiments}
\label{experiments}
We empirically validate the performance of PCNs against their overparameterized counterparts. We consider several architectures including VGG-style CNNs~\cite{simonyan_2014} and ResNets~\cite{he_2016} over CIFAR-10~\cite{krizhevsky_2009} and ImageNet~\cite{russakovsky_2015}. To measure performance we consider standard test accuracy metrics. Finally, we evaluate the time and energy improvements that PCNs introduce.

\newparagraph{General setup}
All experiments were executed over TensorFlow. We implement no special optimizations and use only the high level TensorFlow Keras API. All models are trained from scratch. Network architectures and training details are included in the supplementary material. The CIFAR-10 data set consists of 50k train and 10k test 32x32 images split into 10 classes while ImageNet contains 1.28M train and 50k validation images across 1k classes. 

\newparagraph{Key takeaways}
We summarize end-to-end results in Table \ref{tab:results_summary}. Although PCNs perform compression early in training, they consistently reduce the number of weights in overparameterized networks yet sacrifice little, if any, accuracy drop. Our method extends from the comparatively simple data set CIFAR-10 to the challenging ImageNet data set. For the former, we achieve comparable compression and accuracy improvements to winning tickets found in~\cite{frankle_2019}, but require no iterative train-prune procedure. For the latter, in one case we observe a reduction in parameters by $5.28\times$, while in the other case we observe improvements to end-model accuracy.    

\begin{table}
\small
    \caption{Summary of results}
    \vspace{7pt}
    \label{tab:results_summary}
    \centering
    \begin{tabular}{l l l l l l}
        \toprule
        Data Set & Network & Trainable Parameters & Accuracy & PCN Parameters & PCN Accuracy\\
        \midrule
        CIFAR-10 & Conv4 & 2,425,930 & 75.16 & 101,810 & \textbf{77.25}\\
        CIFAR-10 & WideResNet-20 & 4,331,978 & 93.60 & 1,323,850 & 93.52\\
        ImageNet & VGG-19 & 143,667,240 & 70.99 & 27,201,576 & 70.27\\
        ImageNet & WideResNet-50 & 98,004,072 & 77.52 & 62,375,016 & \textbf{77.64}\\
        \bottomrule
    \end{tabular}
\end{table}

\begin{wrapfigure}{r}{0.4\linewidth}
    \vspace{-3.5pt}
    \centering
    \includegraphics[width=0.35\textwidth]{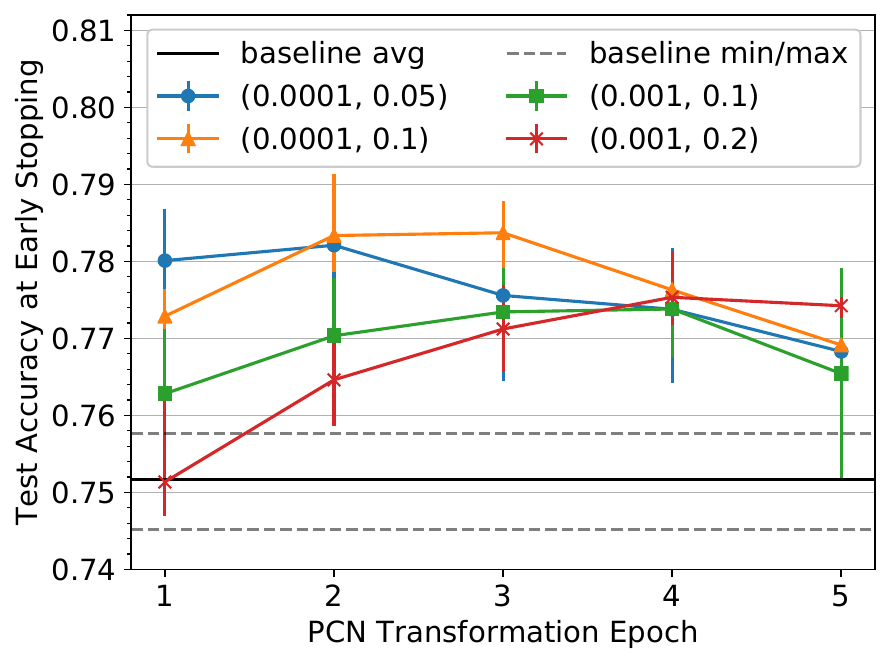}
    \caption{Robustness of PCNs.}
    \vspace{-12pt}
    \label{fig:all_layer}
\end{wrapfigure}
\newparagraph{Robustness of PCN transformations}
We evaluate the sensitivity of PCNs with respect to 1) the transformation epoch and 2) the variance threshold for the input-based transformation.
For this experiment, we use the Conv4 network~\cite{frankle_2019}, a VGG-style CNN adapted to CIFAR-10. It contains four convolution layers followed by three dense layers. In Figure \ref{fig:all_layer} we plot PCN test accuracy at early stopping\footnote{Early stopping is defined here as the epoch of peak validation accuracy.} versus transformation epoch using four different threshold configurations. Variance cutoffs are represented using tuples corresponding to (threshold for convolution layers, threshold for dense layers). The baseline accuracy, i.e. training the parent Conv4 network, is shown as a horizontal line.

We find that training PCNs is robust to both the transformation epoch and the variance thresholds; All configurations in Figure \ref{fig:all_layer} exceed the baseline. In the best case, the Conv4-PCN improves upon the Conv4 accuracy by 3\%. Furthermore, we see that the transformation can be performed early---in this case after one epoch. For reference, training requires 5-12 epochs.

\newparagraph{Origins of accuracy improvement}
To understand why Conv4-PCNs train to a higher accuracy than the baseline, we performed a set of ablation experiments. Details are in the supplementary material. When converting to the PCN, we transformed \textit{all layers but one}. 
Omitting the transformation for the first dense layer causes the accuracy to return to baseline levels, but this drop does not occur when excluding any other layer.
We conclude that by compressing the first dense layer---a layer which contains 86\% of the original weights---the Conv4-PCN helps to regularize and prevent overfitting. 

\newparagraph{Effect of network width}
We evaluate the hypothesis that wide networks improve generalization and that PCNs can help reduce the number of parameters early in training without sacrificing end-model accuracy. On both CIFAR-10 and ImageNet, we trained wide ResNet architectures---networks identical to the original ResNets~\cite{he_2016}, but with more filters at each layer. We compare these networks and their corresponding PCNs with conventional deep ResNet models. Results are shown in Tables \ref{tab:cifar10} and \ref{tab:imagenet}. For CIFAR-10 we show three different PCNs with increasing amount of compression and include three-run max/min for WideResNet-20 and WideResNet-20-PCN0.

We find that wide networks improve accuracy more than deep networks. We also find that PCNs can reduce the number of parameters in wide models while retaining their generalization performance. On CIFAR-10, both PCN0 and PCN1 exceed the accuracy of a deep ResNet-110 with less parameters. On ImageNet, our PCN outperforms a ResNet-152 with a comparable number of trainable weights. 

Additionally, Tables \ref{tab:cifar10} and \ref{tab:imagenet} show that wide networks can be converted into PCNs early in training. The transformation occurs at epoch 15 out of 182 on CIFAR-10 and epoch 16 out of 90 on ImageNet.

\begin{table}
\small
    \caption{Principal Component Networks trained on CIFAR-10}
    \vspace{7pt}
    \label{tab:cifar10}
    \centering
    \begin{tabular}{l l l l l}
        \toprule
        Network Name & Total Params & Trainable Params & Transformation & Test Acc. (Epoch 182) \\
        \midrule
        ResNet-20 & 274,362 & 272,762 & - & 91.03 \\
        ResNet-110 & 1,739,322 & 1,731,002 & - & 92.94 \\
        \cmidrule(r){1-1}
        WideResNet-20 & 4,338,378 & 4,331,978 & - & \textbf{93.60} \; (+0.07, -0.08) \\
        \cmidrule(r){1-1}
        WideResNet-20-PCN0 & 1,443,018 & 1,323,850 & after epoch 15 & \textbf{93.52} \; (+0.10, -0.07)\\
        WideResNet-20-PCN1 & 1,223,114 & 1,094,154 & after epoch 15 & 93.23\\
        WideResNet-20-PCN2 & 541,834 & 473,802 & after epoch 15 & 92.50 \\
        \bottomrule
    \end{tabular}
\end{table}

\begin{table}
\small
    \caption{Principal Component Networks trained on ImageNet}
    \vspace{7pt}
    \label{tab:imagenet}
    \centering
    \begin{tabular}{l l l l p{33mm}}
        \toprule
        Network Name & Total Params & Trainable Params & Transformation & Center Crop Top-1 (Top-5)\newline Val Acc. at Epoch 90\\
        \midrule
        ResNet-50 & 25,610,152 & 25,557,032 & - & 75.60 (92.78) \\
        ResNet-152\footnotemark & 60,344,232 & 60,192,808 & - & 77.00 (93.3) \\
        \cmidrule(r){1-1}
        WideResNet-50 & 98,110,312 & 98,004,072 & - & \textbf{77.52} (93.92) \\
        \cmidrule(r){1-1}
        WideResNet-50-PCN & 71,936,872 & 62,375,016 & after epoch 16 & \textbf{77.64} (93.85) \\
        \bottomrule
    \end{tabular}
\end{table}
\footnotetext{Accuracy reported from https://github.com/kaiminghe/deep-residual-networks.}

\newparagraph{Performance: training time and energy}
We show that by reducing parameter counts early in training, PCNs have the potential to significantly lower the time and energy required to learn high accuracy models. We measure energy usage by integrating \texttt{nvidia-smi} power measurements.

Before discussing results we note the following: As our current implementation uses high level TensorFlow, we are unable to take advantage of optimized GPU kernels at this time. A standard convolution layer fuses padding together with compute, reading a batch of data from memory only once. In contrast, our current implementation pads,\footnote{When transforming a convolution layer, padding is no longer zero.} then multiplies by $U$, then performs convolution, a process which requires three memory I/O operations. For best performance a new GPU kernel is required which fuses all three together. For results below, we simulate fused padding by using the existing convolution kernels, but currently multiply by $U$ separately.

Without an optimized GPU kernel, we already see performance improvements for training. 
Our WideResNet-20-PCN1 on CIFAR-10 trains using 16\% less total energy per epoch after the transformation compared to the full WideResNet-20 and 28\% less energy than ResNet-110 while training 49\% faster per epoch. 
On ImageNet, our \emph{unoptimized} PCN reduces the computational cost, but not the accuracy, of a WideResNet-50, allowing us to exceed the generalization performance of an \emph{optimized} ResNet-152 with the same end-end training time.
We expect fused multiplication with $U$ will significantly improve the energy consumption and training time benefits of PCNs. 

\section{Related Work}
\label{related_work}
A large body of recent work has focused on reducing the number of parameters and associated computational cost of overparameterized neural networks. To benefit model inference, pruning \cite{lecun_1990, hassibi_1993, han_2015, li_2017, hu_2016, srinivas_2015a, yang_2017, he_2017}, distillation \cite{ba_2014, hinton_2015}, quantization \cite{gong_2014}, and weight decomposition \cite{jaderberg_2014, lebedev_2014, kim_2015, denton_2014, zhang_2015, yaguchi_2019} are performed after training. Specifically engineered networks \cite{iandola_2016, howard_2017}, binary weights \cite{zhuang_2019}, and low-rank \cite{denil_2013} architectures help lessen model size from the start, while \cite{srinivas_2017, louizos_2017, srinivas_2016, molchanov_2017, neklyudov_2017, srinivas_2015b, alvarez_2017, xu_2018} aim to sparsify networks during the learning process. Additional works try to identify subnetworks contained in large models \cite{frankle_2019, ramanujan_2019}. We discuss works related to our paper.

\newparagraph{After training}
The compression works of \cite{luo_2017, garg_2020} relate the most to the transformations we use in Section \ref{pcns}. \citet{luo_2017} prune a subset of filters at layer $i$ based on how accurately layer $i+1$ can produce its output. In contrast, we prune a subset of filters based on how accurately the next layer can write its input in the high variance PCA subspace and we execute the transformation early in training. \citet{garg_2020} use PCA on the hidden layer activations to determine the optimal width of each layer and depth of a pre-trained network. They then retrain a new network with optimal dimensions from scratch. We show how to use PCA during training to dynamically reduce network size.

\newparagraph{Before training}
\citet{denil_2013} represent weight matrices using products of low-rank factors. They fix one factor to contain the layer basis and train the other. To construct the basis vectors, they propose to use prior knowledge or to pre-training layers as autoencoders and use the empirical covariance for kernel ridge regression. While PCNs similarly represent layer weight matrices using a fixed basis factor, together with a trainable matrix, the key difference is that we do not impose this structure at the start. Instead we allow the first training epochs of an overparameterized model to identify an appropriate high variance basis of hidden activations and use PCA to find it. We then directly use this basis and rewrite the weight matrix in the corresponding subspace.

\newparagraph{Subnetwork identification}
\citet{frankle_2019} show the existence of small subnetworks embedded in overparameterized models that, when trained in isolation, meet or exceed the accuracy of the full model. They introduce an iterative train-prune-repeat method to identify subnetworks. In place of subnetworks, we show how to find smaller, transformed versions of base networks, and we show how to do this early in a single training procedure. Eliminating the iterative train-prune loop allows us to scale to large data sets such as ImageNet. \citet{ramanujan_2019} find a subset of random weights in an overparameterized model which achieve good generalization performance without any modification, but these randomly weighted subnetworks do not fully match the accuracy of training the base model. In contrast, our goal is to train more efficient networks without sacrificing accuracy.

\section{Conclusions}
\label{conclusions}
In this paper, we show that early in the training process, overparameterized networks can be transformed into smaller versions which train to the same accuracy as the full model. We focus on hidden layer activations rather than model weights, and observe that they exist primarily in subspaces an order of magnitude smaller than the actual network width. This motivates us to introduce Principal Component Networks, architectures which represent layer weights using the high variance subspaces of their input and output activations. We experimentally validate the ability of PCNs to reduce parameter counts early, to train to the same end-model accuracy as overparameterized models, and to reduce the computational costs of learning. In future work, we plan to extend PCNs to Transformers.

\section*{Broader Impact}
This work introduces \emph{a novel resource-efficient deep learning architecture and training procedure}. Principal Component Networks (PCNs) show that by combining statistical procedures such as Principal Component Analysis together with overparameterized neural networks, we can reduce the time and energy required to train high accuracy models. We demonstrate that PCNs are applicable to a diverse set of neural networks, including traditional dense neural networks, convolutional neural networks, and residual neural networks. While our evaluation focuses primarily on computer vision benchmarks, PCNs have immediate implications for machine leaning applications that utilize the aforementioned models. We also see opportunities in using PCNs to develop more efficient attention-based architectures, including Transformers, thus obtaining resource-efficient networks for natural language processing.

In a broader sense, our work on PCNs falls under the call for Green AI. Deep neural networks are known to be energy intensive. New research estimates the carbon footprint of end-to-end development of \textit{a single model} can be equivalent to five times the lifetime emissions of an average car~\cite{strubell_2019}. Even with known architecture and hyperparameters, training Google's BERT language model~\cite{devlin_2018} emits roughly the carbon equivalent of a flight from San Francisco to New York~\cite{strubell_2019}. Clearly, reducing the cost to learn high accuracy neural networks has the potential for many benefits. Researchers will be able to experiment using cloud computing infrastructure for less money, companies like Microsoft, Google, and Amazon could significantly lower the energy consumption of data centers, and the machine learning community as a whole can reduce its carbon footprint.

In addition to reducing the computational cost associated with training overparameterized models, we are also interested in the generalization performance of PCNs beyond the standard test accuracy metrics considered in the paper. Specifically, some of our questions include: Are PCNs more robust to, or prone to, overfitting data sets? Do the high variance PCA subspaces central to the PCN architecture correspond to transferable features, or merely statistical noise? These questions have implications to the performance of PCNs in the presence of adversarial examples, biases in the training data, or distribution shift, and could affect downstream applications. Many conventional architectures are susceptible to even minute natural variations in the data distribution~\cite{recht_2018} and we encourage further research and understanding of PCNs in these settings.

We are excited about the potential for PCNs to reduce training costs across machine learning and plan to improve them by developing optimized GPU kernels and by extending their applicability to state-of-the-art natural language models. We encourage further effort to develop new algorithms and hardware that allow for more energy efficient deep neural networks.

\newpage

\bibliographystyle{abbrvnat}
\bibliography{ms}

\begin{thebibliography}{53}
\providecommand{\natexlab}[1]{#1}
\providecommand{\url}[1]{\texttt{#1}}
\expandafter\ifx\csname urlstyle\endcsname\relax
  \providecommand{\doi}[1]{doi: #1}\else
  \providecommand{\doi}{doi: \begingroup \urlstyle{rm}\Url}\fi

\bibitem[Alvarez and Salzmann(2017)]{alvarez_2017}
J.~M. Alvarez and M.~Salzmann.
\newblock Compression-aware training of deep networks.
\newblock In \emph{Advances in Neural Information Processing Systems}, pages
  856--867, 2017.

\bibitem[Ba and Caruana(2014)]{ba_2014}
J.~Ba and R.~Caruana.
\newblock Do deep nets really need to be deep?
\newblock In \emph{Advances in neural information processing systems}, pages
  2654--2662, 2014.

\bibitem[Blalock et~al.(2020)Blalock, Ortiz, Frankle, and Guttag]{blalock_2020}
D.~Blalock, J.~J.~G. Ortiz, J.~Frankle, and J.~Guttag.
\newblock What is the state of neural network pruning?
\newblock In \emph{Proceedings of Machine Learning and Systems}, 2020.

\bibitem[Chen et~al.(2018)Chen, Wang, Zhao, Papailiopoulos, and
  Koutris]{chen_2018}
L.~Chen, H.~Wang, J.~Zhao, D.~Papailiopoulos, and P.~Koutris.
\newblock The effect of network width on the performance of large-batch
  training.
\newblock In \emph{Advances in Neural Information Processing Systems}, pages
  9302--9309, 2018.

\bibitem[Denil et~al.(2013)Denil, Shakibi, Dinh, Ranzato, and
  de~Freitas]{denil_2013}
M.~Denil, B.~Shakibi, L.~Dinh, M.~A. Ranzato, and N.~de~Freitas.
\newblock Predicting parameters in deep learning.
\newblock In \emph{Advances in Neural Information Processing Systems 26}, pages
  2148--2156, 2013.

\bibitem[Denton et~al.(2014)Denton, Zaremba, Bruna, LeCun, and
  Fergus]{denton_2014}
E.~L. Denton, W.~Zaremba, J.~Bruna, Y.~LeCun, and R.~Fergus.
\newblock Exploiting linear structure within convolutional networks for
  efficient evaluation.
\newblock In \emph{Advances in neural information processing systems}, pages
  1269--1277, 2014.

\bibitem[Devlin et~al.(2018)Devlin, Chang, Lee, and Toutanova]{devlin_2018}
J.~Devlin, M.-W. Chang, K.~Lee, and K.~Toutanova.
\newblock Bert: Pre-training of deep bidirectional transformers for language
  understanding.
\newblock \emph{arXiv preprint arXiv:1810.04805}, 2018.

\bibitem[Frankle and Carbin(2019)]{frankle_2019}
J.~Frankle and M.~Carbin.
\newblock The lottery ticket hypothesis: Finding sparse, trainable neural
  networks.
\newblock In \emph{International Conference on Learning Representations}, 2019.

\bibitem[Garg et~al.(2020)Garg, Panda, and Roy]{garg_2020}
I.~Garg, P.~Panda, and K.~Roy.
\newblock A low effort approach to structured cnn design using pca.
\newblock In \emph{IEEE Access}, volume~8, pages 1347--1360, 2020.

\bibitem[Glorot and Bengio(2010)]{glorot_2010}
X.~Glorot and Y.~Bengio.
\newblock Understanding the difficulty of training deep feedforward neural
  networks.
\newblock In \emph{Proceedings of the thirteenth international conference on
  artificial intelligence and statistics}, pages 249--256, 2010.

\bibitem[Gong et~al.(2014)Gong, Liu, Yang, and Bourdev]{gong_2014}
Y.~Gong, L.~Liu, M.~Yang, and L.~Bourdev.
\newblock Compressing deep convolutional networks using vector quantization.
\newblock \emph{arXiv preprint arXiv:1412.6115}, 2014.

\bibitem[Gunasekar et~al.(2017)Gunasekar, Woodworth, Bhojanapalli, Neyshabur,
  and Srebro]{gunasekar_2017}
S.~Gunasekar, B.~E. Woodworth, S.~Bhojanapalli, B.~Neyshabur, and N.~Srebro.
\newblock Implicit regularization in matrix factorization.
\newblock In \emph{Advances in Neural Information Processing Systems}, pages
  6151--6159, 2017.

\bibitem[Han et~al.(2015)Han, Pool, Tran, and Dally]{han_2015}
S.~Han, J.~Pool, J.~Tran, and W.~J. Dally.
\newblock Learning both weights and connections for efficient neural networks.
\newblock In \emph{Advances in Neural Information Processing Systems 28}, pages
  1135--1143, 2015.

\bibitem[Hassibi and Stork(1993)]{hassibi_1993}
B.~Hassibi and D.~G. Stork.
\newblock Second order derivatives for network pruning: Optimal brain surgeon.
\newblock In \emph{Advances in Neural Information Processing Systems 5}, pages
  164--171, 1993.

\bibitem[He et~al.(2015)He, Zhang, Ren, and Sun]{he_2015}
K.~He, X.~Zhang, S.~Ren, and J.~Sun.
\newblock Delving deep into rectifiers: Surpassing human-level performance on
  imagenet classification.
\newblock In \emph{Proceedings of the IEEE international conference on computer
  vision}, pages 1026--1034, 2015.

\bibitem[He et~al.(2016)He, Zhang, Ren, and Sun]{he_2016}
K.~He, X.~Zhang, S.~Ren, and J.~Sun.
\newblock Deep residual learning for image recognition.
\newblock In \emph{Proceedings of the IEEE conference on computer vision and
  pattern recognition}, pages 770--778, 2016.

\bibitem[He et~al.(2019)He, Zhang, Zhang, Zhang, Xie, and Li]{he_2019}
T.~He, Z.~Zhang, H.~Zhang, Z.~Zhang, J.~Xie, and M.~Li.
\newblock Bag of tricks for image classification with convolutional neural
  networks.
\newblock In \emph{Proceedings of the IEEE Conference on Computer Vision and
  Pattern Recognition}, pages 558--567, 2019.

\bibitem[He et~al.(2017)He, Zhang, and Sun]{he_2017}
Y.~He, X.~Zhang, and J.~Sun.
\newblock Channel pruning for accelerating very deep neural networks.
\newblock In \emph{Proceedings of the IEEE International Conference on Computer
  Vision}, pages 1389--1397, 2017.

\bibitem[Hinton et~al.(2015)Hinton, Vinyals, and Dean]{hinton_2015}
G.~Hinton, O.~Vinyals, and J.~Dean.
\newblock Distilling the knowledge in a neural network.
\newblock \emph{arXiv preprint arXiv:1503.02531}, 2015.

\bibitem[Howard et~al.(2017)Howard, Zhu, Chen, Kalenichenko, Wang, Weyand,
  Andreetto, and Adam]{howard_2017}
A.~G. Howard, M.~Zhu, B.~Chen, D.~Kalenichenko, W.~Wang, T.~Weyand,
  M.~Andreetto, and H.~Adam.
\newblock Mobilenets: Efficient convolutional neural networks for mobile vision
  applications.
\newblock \emph{arXiv preprint arXiv:1704.04861}, 2017.

\bibitem[Hu et~al.(2016)Hu, Peng, Tai, and Tang]{hu_2016}
H.~Hu, R.~Peng, Y.-W. Tai, and C.-K. Tang.
\newblock Network trimming: A data-driven neuron pruning approach towards
  efficient deep architectures.
\newblock \emph{arXiv preprint arXiv:1607.03250}, 2016.

\bibitem[Iandola et~al.(2016)Iandola, Han, Moskewicz, Ashraf, Dally, and
  Keutzer]{iandola_2016}
F.~N. Iandola, S.~Han, M.~W. Moskewicz, K.~Ashraf, W.~J. Dally, and K.~Keutzer.
\newblock Squeezenet: Alexnet-level accuracy with 50x fewer parameters and< 0.5
  mb model size.
\newblock \emph{arXiv preprint arXiv:1602.07360}, 2016.

\bibitem[Ioffe and Szegedy(2015)]{ioffe_2015}
S.~Ioffe and C.~Szegedy.
\newblock Batch normalization: Accelerating deep network training by reducing
  internal covariate shift.
\newblock \emph{arXiv preprint arXiv:1502.03167}, 2015.

\bibitem[Jaderberg et~al.(2014)Jaderberg, Vedaldi, and
  Zisserman]{jaderberg_2014}
M.~Jaderberg, A.~Vedaldi, and A.~Zisserman.
\newblock Speeding up convolutional neural networks with low rank expansions.
\newblock \emph{arXiv preprint arXiv:1405.3866}, 2014.

\bibitem[Kim et~al.(2015)Kim, Park, Yoo, Choi, Yang, and Shin]{kim_2015}
Y.-D. Kim, E.~Park, S.~Yoo, T.~Choi, L.~Yang, and D.~Shin.
\newblock Compression of deep convolutional neural networks for fast and low
  power mobile applications.
\newblock \emph{arXiv preprint arXiv:1511.06530}, 2015.

\bibitem[Kingma and Ba(2014)]{kingma_2014}
D.~P. Kingma and J.~Ba.
\newblock Adam: A method for stochastic optimization.
\newblock \emph{arXiv preprint arXiv:1412.6980}, 2014.

\bibitem[Krizhevsky et~al.(2009)Krizhevsky, Hinton, et~al.]{krizhevsky_2009}
A.~Krizhevsky, G.~Hinton, et~al.
\newblock Learning multiple layers of features from tiny images.
\newblock \emph{University of Toronto}, 2009.

\bibitem[Lebedev et~al.(2014)Lebedev, Ganin, Rakhuba, Oseledets, and
  Lempitsky]{lebedev_2014}
V.~Lebedev, Y.~Ganin, M.~Rakhuba, I.~Oseledets, and V.~Lempitsky.
\newblock Speeding-up convolutional neural networks using fine-tuned
  cp-decomposition.
\newblock \emph{arXiv preprint arXiv:1412.6553}, 2014.

\bibitem[LeCun et~al.(1990)LeCun, Denker, and Solla]{lecun_1990}
Y.~LeCun, J.~S. Denker, and S.~A. Solla.
\newblock Optimal brain damage.
\newblock In \emph{Advances in Neural Information Processing Systems 2}, pages
  598--605, 1990.

\bibitem[LeCun et~al.(1998)LeCun, Bottou, Bengio, and Haffner]{lecun_1998}
Y.~LeCun, L.~Bottou, Y.~Bengio, and P.~Haffner.
\newblock Gradient-based learning applied to document recognition.
\newblock \emph{Proceedings of the IEEE}, 86\penalty0 (11):\penalty0
  2278--2324, 1998.

\bibitem[Li et~al.(2017)Li, Kadav, Durdanovic, Samet, and Graf]{li_2017}
H.~Li, A.~Kadav, I.~Durdanovic, H.~Samet, and H.~P. Graf.
\newblock Pruning filters for efficient convnets.
\newblock In \emph{International Conference on Learning Representations}, 2017.

\bibitem[Li et~al.(2020)Li, Wallace, Shen, Lin, Keutzer, Klein, and
  Gonzalez]{li_2020}
Z.~Li, E.~Wallace, S.~Shen, K.~Lin, K.~Keutzer, D.~Klein, and J.~E. Gonzalez.
\newblock Train large, then compress: Rethinking model size for efficient
  training and inference of transformers.
\newblock \emph{arXiv preprint arXiv:2002.11794}, 2020.

\bibitem[Loshchilov and Hutter(2017)]{loshchilov_2017}
I.~Loshchilov and F.~Hutter.
\newblock Decoupled weight decay regularization.
\newblock \emph{arXiv preprint arXiv:1711.05101}, 2017.

\bibitem[Louizos et~al.(2017)Louizos, Welling, and Kingma]{louizos_2017}
C.~Louizos, M.~Welling, and D.~P. Kingma.
\newblock Learning sparse neural networks through $ l\_0 $ regularization.
\newblock \emph{arXiv preprint arXiv:1712.01312}, 2017.

\bibitem[Luo et~al.(2017)Luo, Wu, and Lin]{luo_2017}
J.-H. Luo, J.~Wu, and W.~Lin.
\newblock Thinet: A filter level pruning method for deep neural network
  compression.
\newblock In \emph{Proceedings of the IEEE international conference on computer
  vision}, pages 5058--5066, 2017.

\bibitem[Molchanov et~al.(2017)Molchanov, Ashukha, and Vetrov]{molchanov_2017}
D.~Molchanov, A.~Ashukha, and D.~Vetrov.
\newblock Variational dropout sparsifies deep neural networks.
\newblock In \emph{Proceedings of the 34th International Conference on Machine
  Learning-Volume 70}, pages 2498--2507. JMLR. org, 2017.

\bibitem[Neklyudov et~al.(2017)Neklyudov, Molchanov, Ashukha, and
  Vetrov]{neklyudov_2017}
K.~Neklyudov, D.~Molchanov, A.~Ashukha, and D.~P. Vetrov.
\newblock Structured bayesian pruning via log-normal multiplicative noise.
\newblock In \emph{Advances in Neural Information Processing Systems}, pages
  6775--6784, 2017.

\bibitem[Park et~al.(2019)Park, Sohl-Dickstein, Le, and Smith]{park_2019}
D.~S. Park, J.~Sohl-Dickstein, Q.~V. Le, and S.~L. Smith.
\newblock The effect of network width on stochastic gradient descent and
  generalization: an empirical study.
\newblock \emph{arXiv preprint arXiv:1905.03776}, 2019.

\bibitem[Ramanujan et~al.(2019)Ramanujan, Wortsman, Kembhavi, Farhadi, and
  Rastegari]{ramanujan_2019}
V.~Ramanujan, M.~Wortsman, A.~Kembhavi, A.~Farhadi, and M.~Rastegari.
\newblock What's hidden in a randomly weighted neural network?
\newblock \emph{arXiv preprint arXiv:1911.13299}, 2019.

\bibitem[Recht et~al.(2018)Recht, Roelofs, Schmidt, and Shankar]{recht_2018}
B.~Recht, R.~Roelofs, L.~Schmidt, and V.~Shankar.
\newblock Do cifar-10 classifiers generalize to cifar-10?
\newblock \emph{arXiv preprint arXiv:1806.00451}, 2018.

\bibitem[Russakovsky et~al.(2015)Russakovsky, Deng, Su, Krause, Satheesh, Ma,
  Huang, Karpathy, Khosla, Bernstein, et~al.]{russakovsky_2015}
O.~Russakovsky, J.~Deng, H.~Su, J.~Krause, S.~Satheesh, S.~Ma, Z.~Huang,
  A.~Karpathy, A.~Khosla, M.~Bernstein, et~al.
\newblock Imagenet large scale visual recognition challenge.
\newblock \emph{International journal of computer vision}, 115\penalty0
  (3):\penalty0 211--252, 2015.

\bibitem[Simonyan and Zisserman(2014)]{simonyan_2014}
K.~Simonyan and A.~Zisserman.
\newblock Very deep convolutional networks for large-scale image recognition.
\newblock \emph{arXiv preprint arXiv:1409.1556}, 2014.

\bibitem[Srinivas and Babu(2015{\natexlab{a}})]{srinivas_2015a}
S.~Srinivas and R.~V. Babu.
\newblock Data-free parameter pruning for deep neural networks.
\newblock \emph{arXiv preprint arXiv:1507.06149}, 2015{\natexlab{a}}.

\bibitem[Srinivas and Babu(2015{\natexlab{b}})]{srinivas_2015b}
S.~Srinivas and R.~V. Babu.
\newblock Learning neural network architectures using backpropagation.
\newblock \emph{arXiv preprint arXiv:1511.05497}, 2015{\natexlab{b}}.

\bibitem[Srinivas and Babu(2016)]{srinivas_2016}
S.~Srinivas and R.~V. Babu.
\newblock Generalized dropout.
\newblock \emph{arXiv preprint arXiv:1611.06791}, 2016.

\bibitem[Srinivas et~al.(2017)Srinivas, Subramanya, and
  Venkatesh~Babu]{srinivas_2017}
S.~Srinivas, A.~Subramanya, and R.~Venkatesh~Babu.
\newblock Training sparse neural networks.
\newblock In \emph{Proceedings of the IEEE Conference on Computer Vision and
  Pattern Recognition Workshops}, pages 138--145, 2017.

\bibitem[Srivastava et~al.(2014)Srivastava, Hinton, Krizhevsky, Sutskever, and
  Salakhutdinov]{srivastava_2014}
N.~Srivastava, G.~Hinton, A.~Krizhevsky, I.~Sutskever, and R.~Salakhutdinov.
\newblock Dropout: a simple way to prevent neural networks from overfitting.
\newblock \emph{The journal of machine learning research}, 15\penalty0
  (1):\penalty0 1929--1958, 2014.

\bibitem[Strubell et~al.(2019)Strubell, Ganesh, and McCallum]{strubell_2019}
E.~Strubell, A.~Ganesh, and A.~McCallum.
\newblock Energy and policy considerations for deep learning in nlp.
\newblock \emph{arXiv preprint arXiv:1906.02243}, 2019.

\bibitem[Xu et~al.(2018)Xu, Li, Zhang, Wen, Wang, Qi, Chen, Lin, and
  Xiong]{xu_2018}
Y.~Xu, Y.~Li, S.~Zhang, W.~Wen, B.~Wang, Y.~Qi, Y.~Chen, W.~Lin, and H.~Xiong.
\newblock Trained rank pruning for efficient deep neural networks.
\newblock \emph{arXiv preprint arXiv:1812.02402}, 2018.

\bibitem[Yaguchi et~al.(2019)Yaguchi, Suzuki, Nitta, Sakata, and
  Tanizawa]{yaguchi_2019}
A.~Yaguchi, T.~Suzuki, S.~Nitta, Y.~Sakata, and A.~Tanizawa.
\newblock Scalable deep neural networks via low-rank matrix factorization.
\newblock \emph{arXiv preprint arXiv:1910.13141}, 2019.

\bibitem[Yang et~al.(2017)Yang, Chen, and Sze]{yang_2017}
T.-J. Yang, Y.-H. Chen, and V.~Sze.
\newblock Designing energy-efficient convolutional neural networks using
  energy-aware pruning.
\newblock In \emph{Proceedings of the IEEE Conference on Computer Vision and
  Pattern Recognition}, pages 5687--5695, 2017.

\bibitem[Zhang et~al.(2015)Zhang, Zou, Ming, He, and Sun]{zhang_2015}
X.~Zhang, J.~Zou, X.~Ming, K.~He, and J.~Sun.
\newblock Efficient and accurate approximations of nonlinear convolutional
  networks.
\newblock In \emph{Proceedings of the IEEE Conference on Computer Vision and
  pattern Recognition}, pages 1984--1992, 2015.

\bibitem[Zhuang et~al.(2019)Zhuang, Shen, Tan, Liu, and Reid]{zhuang_2019}
B.~Zhuang, C.~Shen, M.~Tan, L.~Liu, and I.~Reid.
\newblock Structured binary neural networks for accurate image classification
  and semantic segmentation.
\newblock In \emph{Proceedings of the IEEE Conference on Computer Vision and
  Pattern Recognition}, pages 413--422, 2019.

\end{thebibliography}

\newpage
\appendix
\section*{Appendix}

\section{Transformation order}
We discuss technical details regarding the order of the input- and output-based transformations.

\newparagraph{1.}
We calculate the PCA basis ($U$) for each layer before performing any transformations. That is, we use the original network's hidden activations rather than first transforming layer $i$ and then using its approximate outputs to compute PCA at layer $i+1$. This choice allows us to calculate PCA at each layer using a forward pass through the original network.

\newparagraph{2.}
Given $U^{i+1}$, the order of the input and output-based transformation at layer $i$ does not matter. Pruning columns of $W^i$ and entries of $\mathbf{b}^i$ then transforming to $\tilde{W}^i$ and $\tilde{\mathbf{b}}^i$ is the same as first transforming to $\tilde{W}^i$ and $\tilde{\mathbf{b}}^i$ and then pruning these transformed variables.

\newparagraph{3.}
The order of the output-based transformation at layer $i$ and the input-based transformation at layer $i+1$ does matter. Suppose we already performed the input-based transformation at layer $i$, and we now plan to perform the output-based transformation at layer $i$. As described in the main body of the paper, to do so we 1) find the subset $S \subseteq [1 \dots n]$ with highest row-wise $L_1$ norm in $U^{i+1}$ and 2) prune columns of $\tilde{W}^i$, entries of $\tilde{\mathbf{b}}^i$, entries of $\boldsymbol{\mu}^{i+1}$, and rows of $U^{i+1}$ with indices in $S^C$. The question is then: do we use the truncated $\boldsymbol{\mu}^{i+1}$ and $U^{i+1}$ to calculate $\tilde{W}^{i+1}$ and $\tilde{b}^{i+1}$ in the input-based transformation at layer $i+1$? Or do we first calculate $\tilde{W}^{i+1}$ and $\tilde{b}^{i+1}$ and then truncate $\boldsymbol{\mu}^{i+1}$ and $U^{i+1}$? In the first case, we must also remove rows in $W^{i+1}$ with indices in $S^C$ so that dimensions match when performing the transformation.

Performing the input-based transformation at layer $i+1$ before truncating $\boldsymbol{\mu}^{i+1}$ and $U^{i+1}$ according to the output-based transformation at layer $i$ means that $\tilde{W}^{i+1}$ and $\tilde{b}^{i+1}$ are written in the unmodified high variance PCA subspace. On the other hand, outputs of layer $i$ which are no longer part of the network contribute to these directions. Currently, we truncate first, and then perform the input-based transformation at layer $i+1$. In the future, we plan to study the performance of each ordering.

\newparagraph{Summary}
Given a compression configuration $C[i]$ for layer $i$, we perform the transformations as follows:
\begin{align*}
    W^i, \mathbf{b}^i, \boldsymbol{\mu}^{i+1}, U^{i+1}, W^{i+1} &\leftarrow \texttt{OutputTransformation}(W^i, \mathbf{b}^i, \boldsymbol{\mu}^{i+1}, U^{i+1}, W^{i+1}, C[i])\\
    \tilde{W}^i, \tilde{\mathbf{b}}^i &\leftarrow \texttt{InputTransformation}(W^i, \mathbf{b}^i, \boldsymbol{\mu}^{i}, U^{i}).
\end{align*}

\section{Transformations for convolutional layers}
We provide technical details to extend the discussion on transforming convolutional layers presented in the main body of the paper.

Consider a convolutional layer:
\begin{align*}
    H^{i+1}_{h' \times w' \times n} = \sigma (H^{i}_{h \times w \times m} * W^{i}_{k_1 \times k_2 \times m \times n} + \mathbf{b}^{i}_n)
\end{align*}
where $h$ and $w$ describe the size of the input data, $m$ is the number of input filters, $k_1 \times k_2$ is the kernel size, and $n$ is the number of output filters.

\newparagraph{PCA for images} Given a batch of $N$ input images to layer $i$, compute PCA as follows:
\begin{enumerate}
    \item $H^i_{N' \times m} = \texttt{Flatten}(H^i_{N \times h \times w \times m})$
    \item $\boldsymbol{\mu}^i_m, \mathbf{e}^i_m, V^i_{m \times m} = \texttt{PCA}(H^i_{N' \times m})$
\end{enumerate}

\newparagraph{Input-based transformation}
Transform an input image to the PCA version as so:
\begin{enumerate}
    \item $H^i_{(h \times w) \times m} = \texttt{Flatten}(H^i_{h \times w \times m})$
    \item $\tilde{H}^i_{(h \times w) \times m_e} = (H^i_{(h \times w) \times m} - \boldsymbol{\mu}^i_m)U^i_{m \times m_e}$ (subtraction applies to all rows)
    \item $\tilde{H}^i_{h \times w \times m_e} = \texttt{Reshape}(\tilde{H}^i_{(h \times w) \times m_e})$
\end{enumerate}
You can transform a batch of images all at once by flattening $H^i_{N \times h \times w \times m}$. Note that if the original layer performed zero padding, then the padding should be included in the transformation. Either $H^i_{h \times w \times m}$ should be zero padded and then transformed, in which case no additional padding is required when performing convolution, or channel $j$ of $\tilde{H}^i_{h \times w \times m_e}$ should be padded with $(-\boldsymbol{\mu}^i_m U^i_{m \times m_e})[j]$ during convolution.

Transform the weights $W^i$ by:
\begin{enumerate}
    \item $W^{i}_{n \times m \times (k_1 \times k_2)} = \texttt{Reshape}(W^{i}_{k_1 \times k_2 \times m \times n})$
    \item $\tilde{W}^i_{n \times m_e \times (k_1 \times k_2)} = \texttt{BatchMatrixMultiply}((U^i_{m \times m_e})^T, W^{i}_{n \times m \times (k_1 \times k_2)})$
    \item $\tilde{W}^i_{ k_1 \times k_2 \times m_e \times n} = \texttt{Reshape}(\tilde{W}^i_{n \times m_e \times (k_1 \times k_2)})$
\end{enumerate}

Transform the bias $\mathbf{b}^i$ by:
\begin{enumerate}
    \item $W^{i}_{n \times m \times (k_1 \times k_2)} = \texttt{Reshape}(W^{i}_{k_1 \times k_2 \times m \times n})$
    \item $T_{n \times (k_1 \times k_2)} = \texttt{BatchMatrixMultiply}(\boldsymbol{\mu}^i_m, W^{i}_{n \times m \times (k_1 \times k_2)})$
    \item $\tilde{\mathbf{b}}^i_n = \mathbf{b}^i_n + \texttt{ReduceSum}_{axis=1}(T_{n \times (k_1 \times k_2)})$
\end{enumerate}

The resulting convolutional layer, transformed based on the input hidden activations, computes an approximate output image:
\begin{align*}
    \hat{H}^{i+1}_{h' \times w' \times n} = \sigma (\tilde{H}^{i}_{h \times w \times m_e} * \tilde{W}^{i}_{k_1 \times k_2 \times m_e \times n} + \tilde{\mathbf{b}}^{i}_n).
\end{align*}
Instead of containing $m$ filters, the input images contain only $m_e$ filters of high variance. Only the weights which act on these filters remain in $\tilde{W}^i$. 

\newparagraph{Output-based transformation}
Just as in the main text for dense layers, we use the $L_1$ norm of row $l$ in $U^{i+1}$ to determine the importance of the $l^{th}$ output filter of layer $i$. Given the subset $S \subseteq [1 \dots n]$ of indices with highest row-wise $L_1$ norm in $U^{i+1}$, we can prune filters from layer $i$ with indices in $S^C$. As before, we must also update $\boldsymbol{\mu}^{i+1}$, $U^{i+1}$, and $W^{i+1}$. We include the dimensions below for clarity on what is pruned in the higher dimensional tensors (assuming we already did the input transformation at layer $i$):
\begin{align*}
    \tilde{W}^{i}_{k_1 \times k_2 \times m_e \times |S|}, \mathbf{b}^i_{|S|}, \boldsymbol{\mu}^{i+1}_{|S|}, U^{i+1}_{|S| \times n_e}, W^{i+1}_{k'_1 \times k'_2 \times |S| \times o}.
\end{align*}

When performing the output-based transformation to a convolutional layer $i$ which is followed by a dense layer in the original network, additional steps need to be taken. In this case, the output of layer $i$, $H^{i+1}_{h' \times w' \times n}$, is flattened to $\mathbf{h}^{i+1}_{(h' \times w' \times n)}$. Thus, $U^{i+1}$ has $(h' \times w' \times n)$ rows instead of $n$ rows. To determine the importance of the $l^{th}$ output filter, we add together the $L_1$ norm of each of the corresponding $(h' \times w')$ rows in $U^{i+1}$. Pruning a single filter then requires removing all corresponding $(h' \times w')$ entries in $\boldsymbol{\mu}^{i+1}$, rows in $U^{i+1}$, and rows in $W^{i+1}$ (which is just a matrix again).

\section{Transformations for ResNets}
The input-based convolutional transformation can be directly applied to all layers in a given ResNet. Here, we discuss the additional constraint we impose to perform the output-based transformation. 

\newpage
\begin{wrapfigure}{r}{0.5\linewidth}
    \centering
    \includegraphics[width=0.45\textwidth]{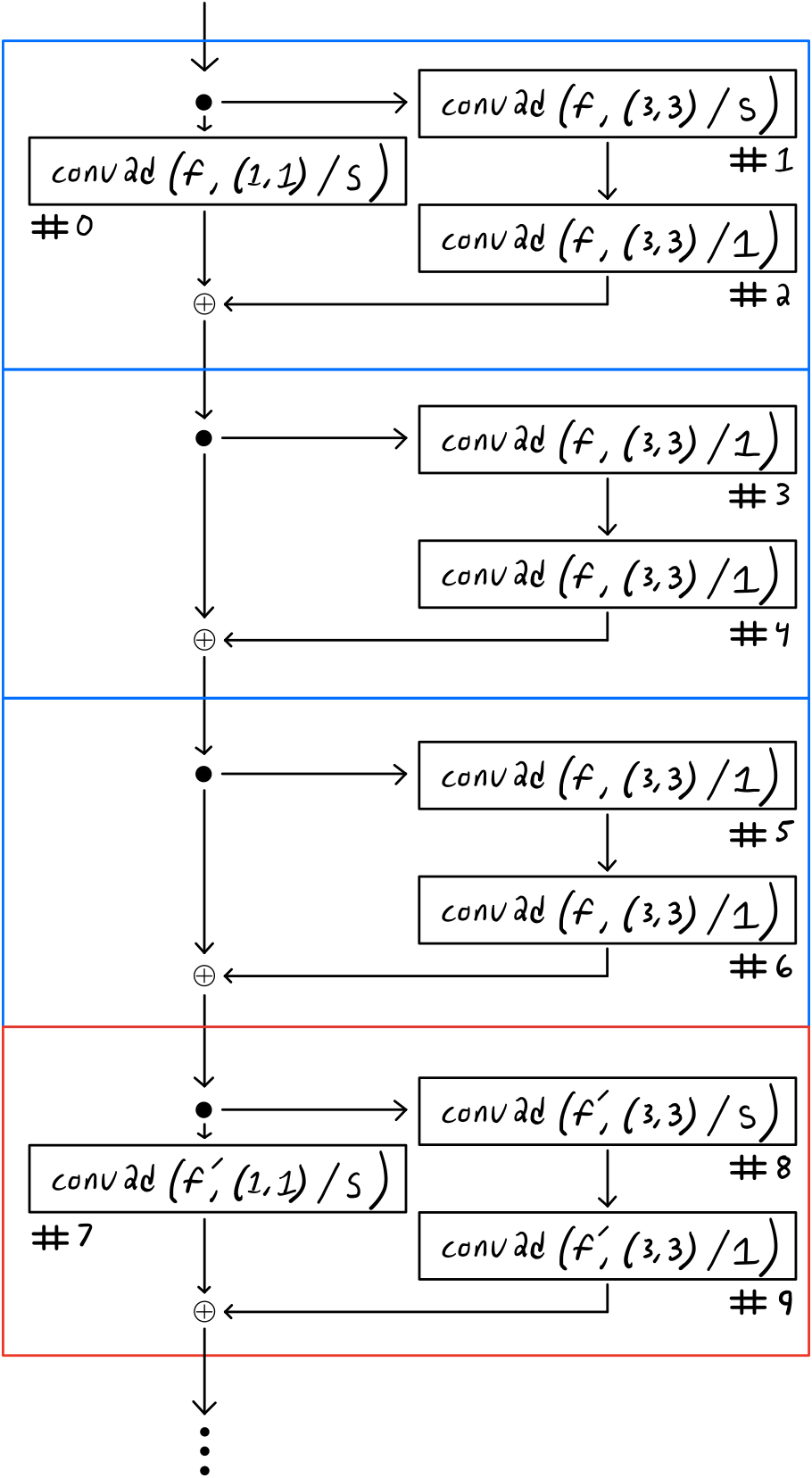}
    \caption{ResNet illustration.}
    \label{fig:resnet_illustration}
    \vspace{-12pt}
\end{wrapfigure}
Consider the partial ResNet diagram shown in Figure \ref{fig:resnet_illustration}. We show a stage with three blocks followed by the first block of the next stage. Convolutional layers are depicted using the tuple (filters, kernel size/strides). Typically, $S=2$ to downsample the image at the beginning of each stage. We omit batch normalization~\cite{ioffe_2015} and activation layers for brevity. All layers $[0, 9]$ can perform the input-based transformation without affecting any other layers. Note that Layers 0 and 1 share the same input activations and thus $U^0=U^1$. The same is true for Layers 7 and 8.

We discuss the output-based transformation for layers in the first stage, i.e. layers $[0, 6]$. First note that Layers 1, 3, and 5 can perform the output-based transformation as if they were in a standard convolutional neural network. Since the output filters of Layers 0, 2, 4, and 6 are added together, we enforce the constraint that if these layers are to prune filters, they must prune the same filters.

To decide which filters to prune, we no longer have only one $U^{i+1}$. Instead, $U^3$, $U^5$, and $U^7=U^8$ all provide information regarding the importance of specific filters. For the $l^{th}$ filter of Layers 0, 2, 4, and 6, we use the average $L_1$ norm of the $l^{th}$ rows in $U^3$, $U^5$, and $U^7$ as the influence measurement. Given this criteria, we can perform the output-based transformation for the even layers in the first stage. Just as $\boldsymbol{\mu}^{i+1}$, $U^{i+1}$, and $W^{i+1}$ are updated in the standard output-based transformation, here we must update $\boldsymbol{\mu}$, $U$, and $W$ for layers 3, 5, 7, and 8.

\section{Transformation overhead}
We discuss the overhead required to convert a network into its corresponding PCN version. In general, the transformation time is negligible compared to the overall training procedure. The primary bottleneck is computing hidden layer activations, which require partial forward passes through the original network. That said, however, only a small fraction of the training data is needed for sufficient PCA statistics. Thus, PCN conversion typically requires much less compute than one training epoch. On CIFAR-10, we transform WideResNet-20 architectures in $\approx 3$s while a single training epoch takes $\approx 17$s. On ImageNet, PCN conversion can be performed an order of magnitude faster than training one epoch. A single GPU transforms a WideResNet-50 in half an hour yet it takes the same time for eight GPUs to execute one epoch.

\section{Network architectures and training details}
In this section, we present details on the architectures and training procedures used in the main body of the paper.

\newparagraph{CIFAR-10 network architectures}
Parent networks for CIFAR-10, before transforming to PCN versions, are given in Table \ref{tab:cifar10_networks}. Convolutional layers use $3 \times 3$ kernels unless otherwise stated. Numbers represent filters/neurons, M represents $2 \times 2$ max pooling, and A represents filter-wise average pooling. Brackets denote residual blocks, with the multiplier out front denoting the number of blocks in the residual stage. All layers use \texttt{ReLU} activation, except the last dense layers where we use \texttt{Softmax}, and all convolutional layers use zero-based ''same'' padding. For ResNet architectures, we use the original versions presented in~\cite{he_2016}. That is, we adopt batch normalization after convolution. Convolution layers in ResNets do not use a bias. While not explicitly necessary in the first stage, for symmetry we always use a projection shortcut ($1\times 1$ convolution) at the first residual block in each stage. Downsampling is performed by stride two convolutions in the first block of stages two and three.

\begin{table}[t]
    \caption{Network architectures for CIFAR-10}
    \vspace{7pt}
    \label{tab:cifar10_networks}
    \centering
    \begin{tabular}{l p{20mm} p{22mm} p{23mm} p{23mm}}
        \toprule
        \textit{Name} & Conv4~\cite{frankle_2019} & ResNet-20~\cite{he_2016} & ResNet-110~\cite{he_2016} & WideResNet-20\\
        \midrule
        \textit{Conv Layers} & 
        64, 64, M\newline128, 128, M & 
        16\newline3x[16, 16]\newline3x[32, 32]\newline3x[64, 64] & 16\newline18x[16, 16]\newline18x[32, 32]\newline18x[64, 64] & 64\newline3x[64, 64]\newline3x[128, 128]\newline3x[256, 256] \\
        \textit{FC Layers} & 
        256, 256, 10 & 
        A, 10 &
        A, 10 &
        A, 10 \\
        \bottomrule
    \end{tabular}
\end{table}

\newparagraph{ImageNet network architectures}
Parent networks for ImageNet are shown in Table \ref{tab:imagenet_networks}. Notation is the same as for CIFAR-10 above. For ResNet architectures, initial convolution layers use $7\times7$ kernels with stride two and initial max pooling layers uses $3\times3$ pooling with stride two. We use bottleneck blocks~\cite{he_2016} where the first and third convolution use $1\times1$ kernels and only the middle layer uses $3\times3$ kernels. As for ResNets on CIFAR-10, we use projection shortcuts at the beginning of every stage. We follow the widely adopted implementation which performs downsampling using stride two convolutions at the $3\times3$ convolutional layers in bottleneck blocks rather than the first $1\times1$ layer. 

\begin{table}[t]
    \caption{Network architectures for ImageNet}
    \vspace{7pt}
    \label{tab:imagenet_networks}
    \centering
    \begin{tabular}{p{14.5mm} p{19mm} p{27mm} p{28mm} p{30mm}}
        \toprule
        \textit{Name} & VGG-19~\cite{simonyan_2014} & ResNet-50~\cite{he_2016} & ResNet-152~\cite{he_2016} & WideResNet-50\\
        \midrule
        \textit{Conv\newline Layers} & 
        2x64, M\newline2x128, M\newline4x256, M\newline4x512, M\newline4x512, M &
        
        64, M\newline3x[64, 64, 256]\newline4x[128, 128, 512]\newline6x[256, 256, 1024]\newline3x[512, 512, 2048] & 
        
        64, M\newline3x[64, 64, 256]\newline8x[128, 128, 512]\newline36x[256, 256, 1024]\newline3x[512, 512, 2048] &
        
        128, M\newline3x[128, 128, 512]\newline4x[256, 256, 1024]\newline6x[512, 512, 2048]\newline3x[1024, 1024, 4096]
        \\
        
        \textit{FC Layers} & 
        2x4096, 1000 & 
        A, 1000 &
        A, 1000 &
        A, 1000 \\
        \bottomrule
    \end{tabular}
\end{table}

\newparagraph{Training Conv4 on CIFAR-10}
For all experiments, we train the parent Conv4 model and the Conv4-PCNs in the same manner. Specifically we use a batch size of 60, Adam~\cite{kingma_2014} optimizer with default TensorFlow learning rate 0.001, cross-entropy loss, and when applicable 5000 random training samples for computing PCA statistics. Since the network and data set are small, we can compute PCA for all required layers using a single forward pass with batch size 5000. We train on a random 45k/5k train/val split (different each run) and train for a maximum of 20 epochs. When applicable, early stopping is defined as the epoch of maximum validation set accuracy. We evaluate on the test set and run each experiment five times. When shown, error bars correspond to five run max/min values, otherwise values correspond to five run mean. We do not use any data set preprocessing. Experiments were run on a machine with one NVIDIA Tesla V100 GPU.

The Conv4-PCN reported in the main text summary of results (with 101,810 trainable parameters) was created using the following compression configuration: \texttt{{conv1:(None, 40), conv2:(20, 50), conv3:(40, 100), conv4:(80, 60), fc1:(50, 90), fc2:(40, 180),\\ output:(30, None)}}. Tuples for each layer represent the effective dimensionality for the input-based transformation and the number of dimensions to keep for the output-based transformation respectively. For this PCN, we perform compression after epoch two.

\newparagraph{Training ResNet on CIFAR-10}
We use data augmentation as reported in~\cite{he_2016}. Namely, "4 pixels are padded on each side, and a $32 \times 32$ crop is randomly sampled from the padded image or its horizontal flip. For testing, we only evaluate the single view of the original $32 \times 32$ image". Additionally, we standardize each channel by subtracting off the mean and dividing by the standard deviation computed over the training data. As for the Conv4 network, we use a 45k/5k train/val split and 5000 samples for compression. Again, a single forward pass suffices for computing PCA statistics. We do not use data augmentation for compression samples. We follow the hyperparameters reported in~\cite{he_2016}: We train for 182 total epochs using a batch size of 128 and an initial learning rate of 0.1 which we drop by a factor of 10 after epochs 91 and 136. For ResNet-110 we use a 0.01 learning rate warm up for the first epoch. We use SGD with momentum 0.9, cross-entropy loss, and $L_2$ regularization coefficient of $5*10^{-5}$ for CNN and dense weights. Note that because TensorFlow incorporates $L_2$ regularization into the total loss, after taking the derivative the coefficient for weight decay becomes $1*10^{-4}$. This matches they weight decay used in different implementations. We do not use weight decay for batch normalization parameters. For CNN layers, we adopt \texttt{he-normal/kaiming-normal} initialization~\cite{he_2015}. We run each experiment three times and report the average test accuracy after epoch 182. Experiments were again run on a machine with one NVIDIA Tesla V100 GPU.

To create the WideResNet-20-PCN0, we use the input-based transformation for all convolutional layers in stages two and three of WideResNet-20 (except the first convolution and projection shortcut in stage two). We use a fixed effective dimensionality of either 32 or 64, corresponding to the input dimension of the respective layer in a ResNet-20. 
For WideResNet-20-PCN1, we use the input-based transformation for all CNN layers in all residual blocks. As for PCN0, we use an effective dimensionality of 16, 32, or 64, the equivalent input dimension of each layer in a ResNet-20.
For WideResNet-20-PCN2, we further compress PCN1 by using the input-based transformation for the dense layer (effective dimensionality 64) and the output-based transformation for all CNN layers in stage three (retaining 64 out of 256 filters). 

\newparagraph{Training VGG on ImageNet}
We adopt the basic training procedure from the original paper~\cite{simonyan_2014}. For train set data augmentation, we randomly sample a $224\times224$ crop from images isotropically rescaled to have smallest side length equal to 256. We include random horizontal flips, but omit random RGB color shift. We subtract the mean RGB values, computed on the train set, from each channel in an image. For testing, we use the center $224\times224$ crop from validation images rescaled to have smallest side 256. We train for 70 total epochs using a batch size of 256 and an initial learning rate of 0.01. We decrease the learning rate by a factor of 10 after epochs 50 and 60. We use $5*10^{-4}$ weight decay decoupled from training loss~\cite{loshchilov_2017}, but multiply by the learning rate to match other implementations. Optimization is done using SGD with momentum 0.9 and cross-entropy loss. We use dropout~\cite{srivastava_2014} with rate 50\% for the first two dense layers. We use default TensorFlow weight initializations (\texttt{glorot-uniform}~\cite{glorot_2010} for CNN layers) and train on AWS p3.16xlarge instances with eight NVIDIA Tesla V100 GPUs. We train networks only once for cost considerations.

We create the PCN presented in the main text summary of results by using the input-based transformation for the first two dense layers in the VGG-19 network. We use an effective dimensionality of 350 and 400 respectively. For PCA statistics, we use the hidden layer activations computed from the center crops of 50,176 (a multiple of the batch size) training images. We randomly sample a different set of images for each layer to increase the amount of training data used to create the PCN. We include dropout with rate 50\% on the two compressed layers. For the network presented in the main text, we perform the transformation after epoch 15 out of 70.

\newparagraph{Training ResNet on ImageNet}
For train set data augmentation, we use the widely adopted procedure described as the baseline in~\cite{he_2019}. Namely, we first randomly crop a region with aspect ratio sampled in [3/4, 4/3] and area randomly sampled in [8\%, 100\%]. This procedure is sometimes referred to as \texttt{RandomResizedCrop}. We then resize the crop to $224\times224$ and perform a random horizontal flip. The brightness, saturation, and hue are randomly adjusted with coefficients drawn from [0.6, 1.4]. We also use PCA color augmentation with coefficients sampled from $N(0, 0.1)$. Finally, we perform channel standardization by subtracting the mean RGB values and dividing by the standard deviations computed across the training data. For validation, we use the center crop of an image isotropically rescaled to have shorter side length 256.

We use the training procedure introduced in the original paper~\cite{he_2016}. We train for 90 total epochs, use a batch size of 256, and an initial learning rate of 0.1. We drop the learning rate by a factor of 10 after epochs 30 and 60. We use SGD with momentum 0.9, cross-entropy loss, and $L_2$ regularization with coefficient $5*10^{-5}$. When multiplied by two in the derivative of the loss function, this $L_2$ penalty equals the $1*10^{-4}$ weight decay used in other implementations. We do not use weight decay for batch normalization parameters. For convolution layers, we use \texttt{he-normal} initialization~\cite{he_2015}. Following common practice, we initialize $\mathbf{\gamma}$ in the final batch normalization of a residual block with zeros instead of ones and use label smoothing with coefficient 0.1. We train on AWS p3.16xlarge instances with eight NVIDIA Tesla V100 GPUs. We train networks only once for cost considerations.

For the WideResNet-50-PCN, we transform all convolutional layers in stage four of a WideResNet-50 using the input-based transformation. We use an effective dimensionality of 512 for transformed layers. To compute PCA statistics, we use the center crops of training images isotropically rescaled to shorter side length 256. We use 50,176 (a multiple of the batch size) random images for each layer. Since the hidden layer activations occupy significant GPU memory, once a batch of images reaches the hidden layer of interest, we downsample using spatial average pooling and use depth vectors from the smaller images for PCA.

\section{Origins of accuracy improvement}
We include the results of the origins of accuracy improvement ablation experiment discussed in the main body of the paper.

\begin{wrapfigure}{r}{0.5\linewidth}
    \centering
    \includegraphics[width=0.45\textwidth]{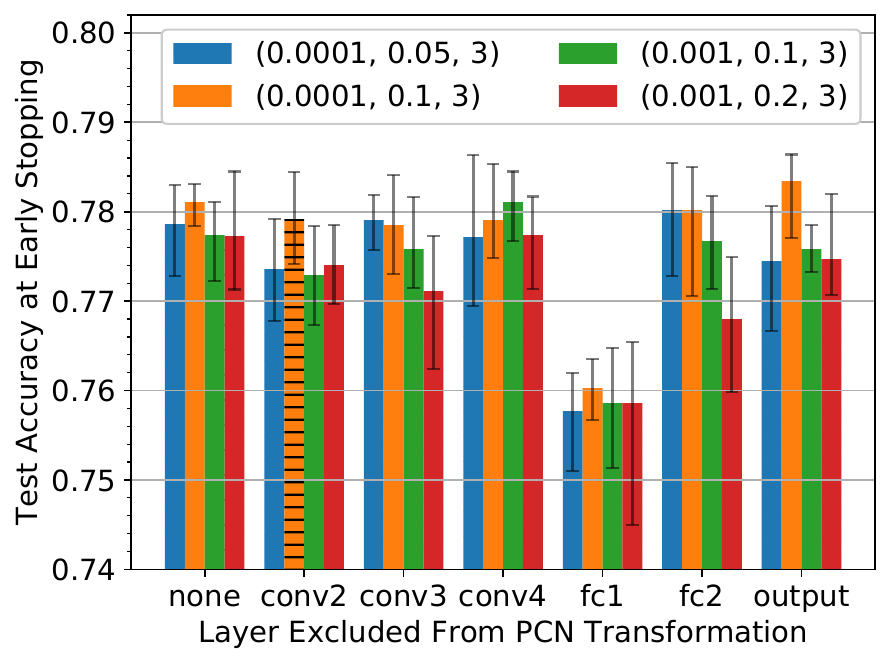}
    \caption{Origins of accuracy improvement.}
    \label{fig:ablation_accuracy}
\end{wrapfigure}
To understand why Conv4-PCNs train to a higher test accuracy than the Conv4 network, we ran a set of ablation experiments. When converting the Conv4 network into the Conv4-PCN, instead of transforming all layers using the input-based transformation, we transformed all layers except one. The layer not transformed is shown on the x-axis of Figure \ref{fig:ablation_accuracy}. We then measured the test accuracy of the resulting PCNs (y-axis). For each PCN, we show four bars labeled by a tuple representing the variance threshold for convolution layers that were compressed, the variance threshold for dense layers that were compressed, and the epoch at which we performed the transformations. 

As discussed in the main body of the paper, omitting the transformation for the first fully connected layer (fc1) causes the accuracy to drop much closer to the baseline (75.16). Omitting the transformation for other layers does not change the PCN accuracy. From this ablation experiment, we conclude that the accuracy improvement of the Conv4-PCN comes from transforming the first dense layer.

\end{document}